\documentclass{article}

\usepackage{microtype}
\usepackage{graphics,graphicx,caption,float,color}
\usepackage{subfigure}
\usepackage{booktabs} 
\usepackage{adjustbox}
\usepackage{multirow}
\usepackage{multicol}
\usepackage{amsmath}
\usepackage{xcolor}
\usepackage{ulem}
\usepackage{amsfonts} 
\usepackage{colortbl}
\usepackage{parskip}
\usepackage{makecell} 
\definecolor{baselinecolor}{gray}{.9}
\newcommand{\baseline}[1]{\cellcolor{baselinecolor}{#1}}
\newcommand{\textred}[1]{\color[HTML]{E92525}{#1}}

\usepackage{hyperref}

\usepackage[accepted]{icml2023}

\usepackage{amsmath}
\usepackage{amssymb}
\usepackage{mathtools}
\usepackage{amsthm}

\usepackage[capitalize,noabbrev]{cleveref}

\theoremstyle{plain}
\newtheorem{theorem}{Theorem}[section]

\theoremstyle{definition}

\newtheorem{assumption}[theorem]{Assumption}
\theoremstyle{remark}

\usepackage[textsize=tiny]{todonotes}

\icmltitlerunning{$\pi$-Tuning: Transferring Multimodal Foundation Models with Optimal Multi-task Interpolation}

\begin{document}

\twocolumn[
\icmltitle{$\pi$-Tuning: Transferring Multimodal Foundation Models \\with Optimal Multi-task Interpolation}



\icmlsetsymbol{equal}{*}

\begin{icmlauthorlist}
\icmlauthor{Chengyue Wu}{hku,tencent}
\icmlauthor{Teng Wang}{hku,tencent}
\icmlauthor{Yixiao Ge}{tencent}
\icmlauthor{Zeyu Lu}{sjtu}
\icmlauthor{Ruisong Zhou}{fudan}
\icmlauthor{Ying Shan}{tencent}
\icmlauthor{Ping Luo}{hku}
\end{icmlauthorlist}

\icmlaffiliation{hku}{Department of Computer Science, The University of Hong Kong}
\icmlaffiliation{tencent}{ARC Lab, Tencent PCG}
\icmlaffiliation{sjtu}{Shanghai Jiao Tong University}
\icmlaffiliation{fudan}{School of Mathematical Sciences, Fudan University}

\icmlcorrespondingauthor{Yixiao Ge}{yixiaoge@tencent.com}

\icmlkeywords{multi-task, transfer learning, interpolation, robustness, multimodal}

\vskip 0.3in
]



\printAffiliationsAndNotice{}  


\begin{abstract}

Foundation models have achieved great advances in multi-task learning with a unified interface of unimodal and multimodal tasks.
However, the potential of such multi-task learners has not been exploited during transfer learning.
In this work, we present a universal parameter-efficient transfer learning method, termed \textbf{P}redict-\textbf{I}nterpolate Tuning ($\pi$-Tuning), for vision, language, and vision-language tasks. It aggregates the parameters of lightweight task-specific experts learned from similar tasks to aid the target downstream task. 
The task similarities are predicted in a unified modality-independent space, yielding a scalable graph to demonstrate task relationships.
%
$\pi$-Tuning has several appealing benefits.
First, it flexibly explores both intra- and inter-modal transferability between similar tasks to improve the accuracy and robustness of transfer learning, especially in data-scarce scenarios.
Second, it offers a systematical solution for transfer learning with multi-task prediction-and-then-interpolation, compatible with diverse types of parameter-efficient experts, such as prompt and adapter.
%
%
Third, an extensive study of task-level mutual benefits on 14 unimodal and 6 multimodal datasets shows that $\pi$-Tuning surpasses fine-tuning and other parameter-efficient transfer learning methods both in full-shot and low-shot regimes. The task graph also enables an in-depth interpretable analysis of task transferability across modalities.
The code will be available at {\small\url{https://github.com/TencentARC/pi-Tuning}.}

\end{abstract}

\begin{figure}[!t]
\vskip 0.2in
\begin{center}
\includegraphics[width=1\columnwidth]{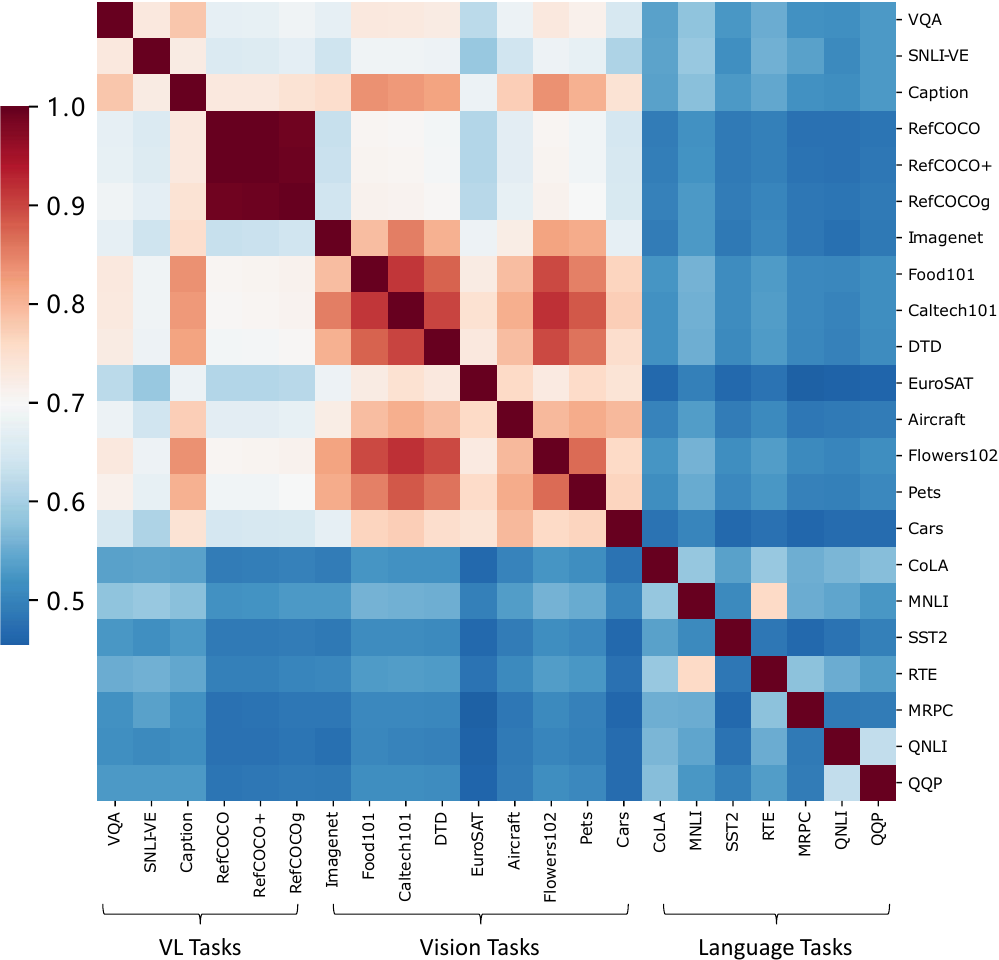}
\caption{Heatmap of the predicted task similarities, composed of both unimodal and multimodal tasks. Vision-language tasks are more similar to vision tasks compared to language tasks. Best viewed in color.}
\label{task_heatmap}
\end{center}

\vskip -0.1in
\end{figure}



\section{Introduction}

With the development of Transformer architectures~\cite{lucas2021vit,devlin2018bert,brown2020gpt3}, foundation models~\cite{cho2021unifying,lu2022unified,wang2022ofa} pre-trained with large-scale data are capable of multiple tasks across modalities in a unified sequence-to-sequence manner, taking one more step toward mimicking the human brain. 
These foundation models are natural multi-task learners with universal representation and I/O interfaces for both unimodal and multimodal tasks. But unfortunately, these properties have not been fully exploited in downstream tasks, as few studies investigated how to properly transfer these models.




In this work, we tackle the problem of transfer learning of multimodal foundation models with unified sequence-to-sequence interfaces.
Most of our experiments are based on OFA~\cite{wang2022ofa}, an open-source model, without loss of generality.
Some previous attempts~\cite{pruksachatkun2020intermediate} empirically observed that pre-finetuning with similar tasks may be beneficial while dissimilar tasks are harmful.
It is intuitive to leverage auxiliary tasks that are similar to the target domain to boost transfer learning.
The measurement of task similarities turns out to be a critical problem.
Rather than the brute-force probing as in \citet{pruksachatkun2020intermediate}, we embed tasks into a unified space with Fisher Information Matrix~\cite{achille2019task2vec}, yielding task graphs across modalities (see Fig.~\ref{task_heatmap}).
We are the first to explore task relationships among computer vision (CV), natural language processing (NLP), and vision-language (VL) tasks.
The graph is computationally efficient and easily scalable for new tasks, which is especially suitable for recent arts that unify increasingly more tasks in one model.

\begin{figure*}[!h]
    \centering
    \includegraphics[width=0.95\linewidth]{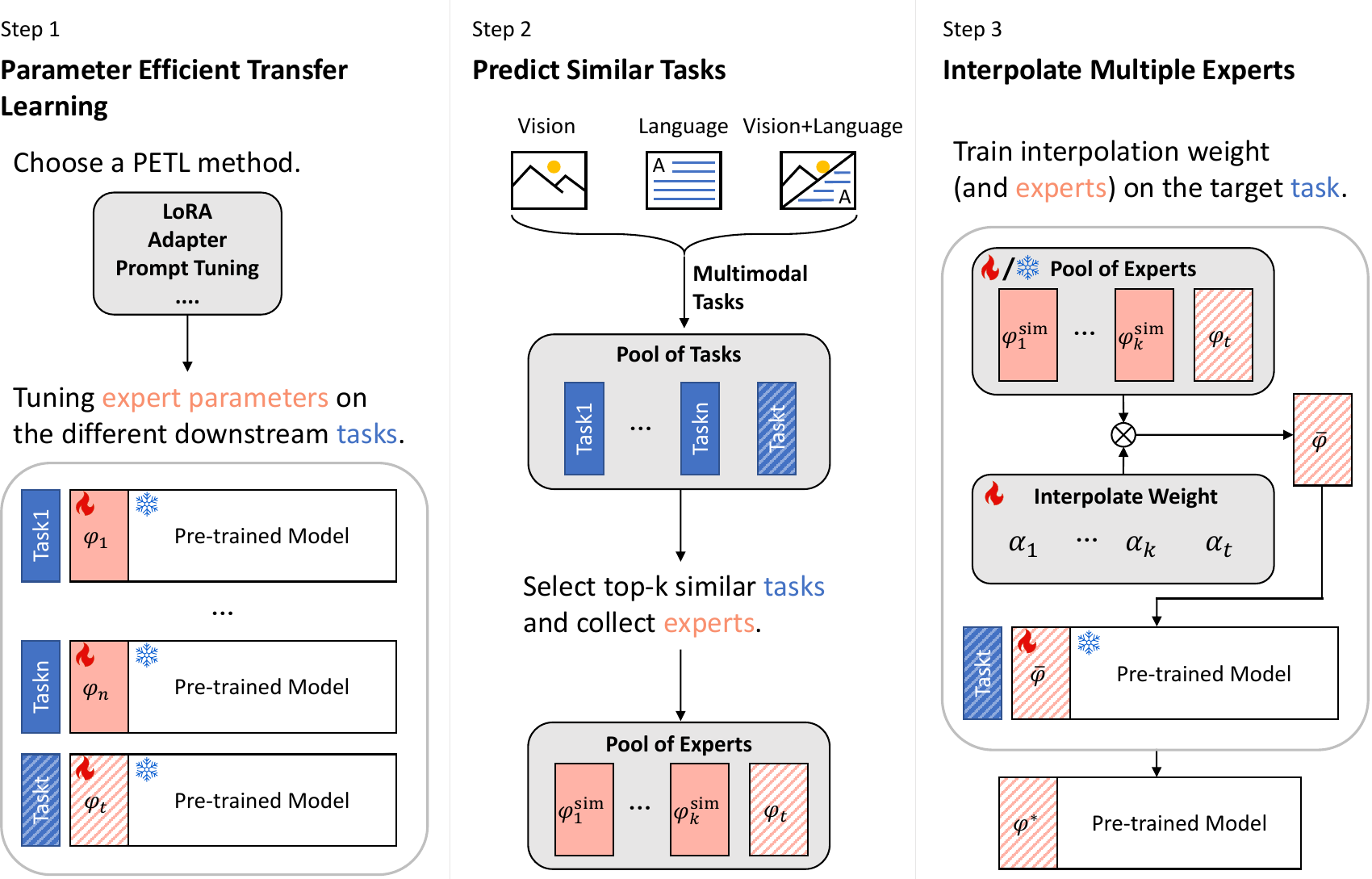}
    \vspace{-0.1cm}
    \caption{Overview of the $\pi$-Tuning method. Step 1 is the traditional parameter-efficient transfer learning (PETL) pipeline. Given the target task, $\pi$-Tuning further adds Step 2 and Step 3 to utilize task relationships to enhance the target expert. Specifically, we find the most similar tasks in a large pool of tasks and interpolate those experts with our target expert on the target task.}
    \vspace{-0.1cm}
    \label{overview}
\end{figure*}

Given the similar tasks predicted from the computed graph, na\"ive multi-task fine-tuning is effective for achieving satisfactory performance but inefficient for training, especially with increasing tasks and model sizes.
Inspired by the parameter-efficient transfer learning methods~\cite{houlsby2019parameter,hu2021lora,li2021prefix} with only a few trainable parameters, we propose \textbf{P}redict-\textbf{I}nterpolate Tuning ($\pi$-Tuning), which interpolates between predicted parameter-efficient experts (\textit{e.g.}, adapter, prompt) for transfer learning.
To be specific, as demonstrated in Fig.~\ref{overview}, the parameter-efficient experts are trained individually on each task before being selected according to the predicted task similarities. The weights of selected experts are then efficiently tuned to be ensembled for the downstream task.
We empirically and theoretically found that those task-specific experts trained on similar tasks may lie in the basin of the loss landscape for the target task, which makes a simple interpolation of different experts possibly yielding strong performance, with no need for elaborately designed and highly parameterized fusion modules~\cite{pfeiffer2020adapterfusion}.

In addition to the newly proposed transfer learning method, several interesting findings were observed by macroscopically analyzing the relationship between CV, NLP, and VL tasks.
1) VL tasks are closer to CV tasks than NLP tasks, interpreting the state-of-the-art performance VL models achieved in vision benchmarks.
2) Image captioning sits in the central position among all the tasks, \textit{i.e.}, being close to many tasks, demonstrating its importance in VL pre-training.
These findings may well inspire future multimodal studies. 

Some pioneer works~\cite{vu2020exploring,poth2021pre,vu2022spot} in NLP attempted to aggregate prompt embeddings from several source tasks to the target domain. However, they focus on limited language tasks and did not systematically come up with the philosophy of task integration, rendering limited scalability for a broader domain.

Our key contributions are three-fold:
\begin{itemize}
    \item We for the first time investigate the task relationships among vision, language, and vision-language tasks in a unified task space. Building upon the task graph, interesting phenomena are demonstrated across modalities.
    \item We introduce a new parameter-efficient transfer learning method, namely $\pi$-Tuning, which effectively and efficiently aggregates unimodal and multimodal knowledge across tasks.
    \item Empirically, we conduct extensive experiments on vision, language, and vision-language downstream tasks to demonstrate the superiority, generalization, and scalability of our approach.
    Theoretically, we indicate the plausibility of similar task interpolation.
\end{itemize}

\label{sec:introduction}

\section{Method}
\subsection{Preliminary}
\label{sec:3.1}

We use a unified sequence-to-sequence (Seq2Seq) model (\textit{e.g.}, OFA~\cite{wang2022ofa}) as it can process heterogeneous data via a universal interface, which is helpful for us to analyze multimodal tasks. 
We specify this model as $y=f_{\theta_0}(x)$, where $\theta_0$ is the pre-treained weight of OFA.
Here, we refer a task to a specific dataset as $\tau_i=(X_i, Y_i)$, where $i\in \{1, 2, \cdots, n\}$ represents the $i$-th task from all of $n$ tasks.
$X_i=\{x^1_i, x^2_i, \cdots, x^{s_i}_i\}$ is the input data, which can be CV, NLP, or VL data and $s_i$ is the dataset size of $\tau_i$. 
$Y_i=\{y^1_i, y^2_i, \cdots, y^{s_i}_i\}$ is the label set, which later will be processed by a verbalizer~\cite{schick2020s} in OFA. 
We utilize parameter-efficient tuning methods, which keep the pre-trained weight $\theta_0$ frozen and only tune extra added parameters $\varphi$ to adapt to multiple downstream tasks. Given a task $\tau_i$, we represent the corresponding optimized parameters as $\varphi_i$.

\textbf{Task Definition.} We have a task set $T=\{\tau_1, \tau_2, \cdots, \tau_n\}$ and the pre-trained model $f_{\theta_0}$. For a specific task $\tau_i$, the objective of traditional parameter-efficient tuning is of the following form:
\begin{align}
\varphi_i \gets \mathop{\arg\min}\limits_{\varphi_i} L(\tau_i; \theta_0, \varphi_i),
\end{align}
where $L$ represents the loss function. In the Seq2Seq model, it is typically the Cross-Entropy (CE) loss.

For the target task $\tau_{t} \in T$, we want to leverage all of the tasks and their corresponding trained parameters $\Phi=\{\varphi_1, \varphi_2, \cdots, \varphi_n\}$ in pursuit of more accurate and robust generalization performance.

\textbf{Fisher Information Matrix (FIM).} 
The Fisher information matrix~\cite{amari1998natural} can indicate the sensitivity of the model towards a small perturbation and the curvature of the loss surface, which can be used as task embedding~\cite{achille2019task2vec}. Formally, it is the expected covariance of the gradients of the log-likelihood for model parameters $\theta$, where $P(x,y)$ represents the data distribution of the task:
\begin{align}
    F_{\theta}^i = \mathop{\mathbb{E}}\limits_{(x,y) \sim P_{\theta}(x,y)}{\nabla_{\theta}{\log P_{\theta}(y|x)} \nabla_{\theta}{\log P_{\theta}(y|x)}^\mathrm{T}}.
\end{align}
Specifically, we use the empirical Fisher to compute task embedding for task $\tau_i$:
\begin{align}
    F_{\theta}^i = \frac{1}{s_i}\sum\limits_{j=1}^{s_i}{[\nabla_{\theta}{\log P_{\theta}(y_i^j|x_i^j)} \nabla_{\theta}{\log P_{\theta}(y_i^j|x_i^j)}^\mathrm{T}]}.
\end{align}


\begin{table*}[t]
\renewcommand\arraystretch{1.5}
\center
\small
\vskip 0.15in
\begin{adjustbox}{max width=1.\textwidth}
\begin{tabular}{lccc|ccc|cc|cc|cccc|cc}
\toprule[1.3pt]
  \multirow{2}*{Method}
  &\multicolumn{3}{c|}{RefCOCO}
  &\multicolumn{3}{c|}{RefCOCO+}
  &\multicolumn{2}{c|}{RefCOCOg}
  &\multicolumn{2}{c|}{SNLI-VE}
  &\multicolumn{4}{c|}{COCO Captions}
  &\multicolumn{2}{c}{VQA}
 
  \\
  & val & testA & testB
  & val & testA & testB
  & val-u & test-u
  & dev & test
  & B@4 & M & C & S
  & test-dev & test-std
  \\
\Xhline{0.5pt}
    \multicolumn{17}{l}{\textit{Previous SOTAs}} \\
    UNITER ~\cite{chen2019uniter}
   & 81.41 & 87.04 & 74.17
   & 75.90 & 81.45 & 66.70
   & 74.86 & 75.77
   & 79.40 & 79.40
   & - & - & - & -
   & 73.80  & 74.00
   \\
   
   VILLA~\cite{gan2020large}
   & 82.39 & 87.48 & 74.84
   & 76.17 & 81.54 & 66.84
   & 76.18 & 76.71
   & 80.20 & 80.00
   & - & - & - & -
   & 74.70  & 74.90
   \\
   
   MDETR~\cite{kamath2021mdetr}
   & 86.75 & 89.58 & 81.41
   & 79.52 & 84.09 & 70.62
   & 81.64 & 80.89
   & 80.90 & 81.20
   & - & - & - & -
   & 77.70  & 77.60
   \\
   
   VL-T5~\cite{cho2021unifying}
   & - & - & -
   & - & - & -
   & 71.20 & 71.30
   & - & -
   & 34.50 & 28.70 & 116.5 & 21.90
   & -  & 70.30
   \\

   UNICORN~\cite{yang2021crossing}
   & 88.29 & 90.42 & 83.06
   & 80.30 & 85.05 & 71.88
   & 83.44 & 83.93
   & - & -
   & 35.80 & 28.40 & 119.10 & 21.50
   & -  & -
    \\
\Xhline{0.5pt}
    \multicolumn{17}{l}{\textit{OFA-Base}} 
    \\
    Finetuning~\cite{wang2022ofa}
    & \uline{88.48} & \uline{90.67} & \uline{83.30}
    & \uline{81.39} & \uline{87.15} & \uline{74.29}
    & \uline{82.29} & \uline{82.31}
    & \uline{89.30} & \uline{89.20}
    & \uline{41.00} & \uline{30.90} & \uline{138.2} & \uline{24.20}
    & \uline{78.00}  & \uline{78.10}
    \\
    BitFit~\cite{zaken2021bitfit}
    & 76.32 & 81.21 & 72.80
    & 67.29 & 74.14 & 59.21
    & 68.79 & 69.61
    & 84.84 & 84.48
    & 39.80 & 30.20  & 134.6 & 23.86
    & 73.03 & 73.26
    \\
    LoRA~\cite{hu2021lora}
    & 81.91 & 85.89 & 76.90
    & 72.29 & 79.22 & 62.28
    & 72.55 & 73.26 
    & 87.83 & 87.93
    & 39.80 & 30.20  & 134.5 & 23.73
    & 75.57 & 75.67
    \\
    Prompt Tuning~\cite{yang2022prompt}
    & 84.53 & 85.21 & 77.36
    & 76.34 & 81.44 & 67.68
    & 75.61 & 76.57
    & 88.18 & 88.59
    & 39.70 & 30.10 & 134.2 & 23.50
    & 74.31 & 74.47
    \\
    Adapter~\cite{houlsby2019parameter}
    & 86.63 & \textbf{90.01} & 81.71 
    & 79.45 & 84.89 & 71.36
    & 79.58 & 80.35
    & 87.90 & 87.67
    & 39.80 & 30.60 & 134.6 & 23.80 
    & 75.59 & 75.94
    \\
    \baseline{$\pi$-Adapter}
    & \baseline{\textbf{86.98}} & \baseline{89.99} & \baseline{\textbf{81.73}}
    & \baseline{\textbf{80.10}} & \baseline{\textbf{85.87}} & \baseline{\textbf{71.38}}
    & \baseline{\textbf{81.72}} & \baseline{\textbf{81.75}}
    & \baseline{\textbf{89.23}} & \baseline{\uline{\textbf{89.40}}}
    & \baseline{\uline{\textbf{41.00}}} & \baseline{\uline{\textbf{30.90}}} & \baseline{\textbf{137.0}} & \baseline{\textbf{23.90}} 
    & \baseline{\textbf{75.88}} & \baseline{\textbf{76.13}}
    \\
\Xhline{0.5pt}
    \multicolumn{17}{l}{\textit{OFA-Large}} 
    \\
    Finetuning
    & 90.05 & \uline{92.93} & 85.26
    & 84.60$^*$  & 89.99$^*$ & 77.71$^*$
    & 85.89 & 86.55
    & \uline{90.36$^*$}  & 89.91$^*$
    & \uline{41.90$^*$} & \uline{31.40$^*$} & \uline{141.8$^*$} & \uline{24.50$^*$}
    & \uline{80.40}  & \uline{80.70}
    \\
    BitFit    
    & 89.61 & 92.20 & 84.91
    & 82.60 & 88.08 & 75.16
    & 84.66 & 84.68
    & 89.70 & 89.42
    & 41.02 & 30.92 & 138.8 & 24.23
    & 78.23  & 78.44
    \\
    LoRA
    & 89.56 & 92.59 & 84.63
    & 83.00 & 88.70 & 75.46
    & 84.48 & 85.01
    & 89.49 & 89.15
    & 41.50 & 31.10  & 140.4 & 24.40
    & 78.20 & 78.16
    \\
    Prompt Tuning
    & 90.05 & 92.31 & 85.59
    & 84.54 & 89.40 & 77.77
    & 85.27 & 85.89
    & 89.19$^*$ & 89.11$^*$
    & 41.60$^*$ & 30.80$^*$ & 140.5$^*$ & 24.30$^*$
    & 78.30  & 78.53
    \\
    Adapter
    & 90.05 & 92.42 & 84.83
    & 84.50 & 89.66 & 77.26
    & 85.48 & 85.88
    & 90.04 & 89.59
    & \textbf{41.80} & 31.30 & 140.6 &  \uline{\textbf{24.50}} 
    & 78.55  & 78.62
    \\
    \baseline{$\pi$-Adapter}
    & \baseline{\uline{\textbf{90.49}}} & \baseline{\uline{\textbf{92.93}}} & \baseline{\uline{\textbf{85.91}}}
    & \baseline{\uline{\textbf{84.92}}} & \baseline{\uline{\textbf{90.03}}} & \baseline{\uline{\textbf{77.91}}}
    & \baseline{\uline{\textbf{86.60}}} & \baseline{\uline{\textbf{86.92}}}
    & \baseline{\textbf{90.16}} & \baseline{\uline{\textbf{90.01}}}
    & \baseline{41.70} & \baseline{\uline{\textbf{31.40}}} & \baseline{\textbf{140.7}} & \baseline{\uline{\textbf{24.50}}}
    & \baseline{\textbf{78.78}} & \baseline{\textbf{78.82}}
    \\
  
\bottomrule[1.3pt]
\end{tabular}
\end{adjustbox}
\caption{Comparison with state-of-the-art (SOTA) PETL and full fine-tuning methods. We report the experimental results on RefCOCO, RefCOCO+, RefCOCOg, SNLI-VE, COCO Image Captioning, and VQA. The overall best result is \uline{underlined} while \textbf{bold} signifies the best among parameter-efficient methods. $^*$ denotes the results of our re-trained models based on  official codes.}
\label{tb:multimodal_results}
\vskip -0.1in
\end{table*}

\subsection{$\pi$-Tuning}
\label{sec:3.2}

Our core hypothesis is that similar tasks help enrich domain-relevant or task-relevant training data, while dissimilar tasks may be harmful to the model to perform well in the target task. Based on this hypothesis, we first retrieve a subset of similar tasks given the target task $\tau_t$ of the task set $T$. Then we combine parameter-efficient experts trained for those similar tasks, which is considered to contain useful knowledge towards corresponding tasks, with the target task expert via interpolation to seek a transfer gain. 
We refer to this approach as Predict-Interpolate Tuning ($\pi$-Tuning for short). Fig.~\ref{overview} shows the overall framework.

\paragraph{Task similarity prediction.}
Since the combination number of the task set $T$ is exponential for $n$, it is computationally expensive to identify the subset $R \subseteq T$ of similar tasks given a target task brutally. Therefore, we try to embed each task into a vector to compute task similarity.  We choose the FIM to extract the task feature as mentioned in Sec. \ref{sec:3.1}. We keep the pre-trained weights $\theta_0$ frozen and tune $\varphi_i$ for each task $\tau_i$ via parameter-efficient tuning.

Next, given task $\tau_i$, we compute the task embedding based on the Fisher of the task expert's weights $\varphi$.
As the dimension of $F_\theta$ is extremely large, we make an approximation by only considering its diagonal entries following the practice in ~\citet{achille2019task2vec}. 
The task embedding $F$ is therefore measured via $F={\rm diag}(F_{\theta})=[F_{\theta_1}, F_{\theta_2}, \cdots, F_{\theta_n}]$, where $\theta_i$ denotes the $i$-th parameter of the expert. Intuitively, the approximation is made under the assumption that correlations between different parametric modules in the expert are not essential.

We compute the cosine similarity between the embedding of the target task $\tau_t$ and the candidate source task. We then rank the similarity score in descending order to select Top-k similar tasks to form the subset $R=\{\tau^{\rm sim}_1, \tau^{\rm sim}_2, \cdots, \tau^{\rm sim}_k\}$. Both the task set $T$ and the similar task subset $R$ are \textit{dynamic}. If there is a new task arrives, we can add it to our task set and compute its task similarity given the target task to check if it is the top-k similar task.

\paragraph{Interpolation of multiple lightweight experts.}
Given the similar task subset $R$, we retrieve the expert subset corresponding to $R$, \textit{i.e.}, $\phi=\{\varphi^{\rm sim}_1,\varphi^{\rm sim}_2,\cdots, \varphi^{\rm sim}_k\} \subseteq \Phi$.


Then we can derive the combined parameters $\Bar{\varphi}$ as follow:
\begin{align}
    \Bar{\varphi} = {\rm softmax}(\alpha)_0\varphi_t +\sum_{i=1}^k {\rm softmax}(\alpha)_i\varphi^{\rm sim}_i.
\end{align}
Then, we may further tune $\Bar{\varphi}$ given the target task $\tau_t$\footnote{We achieve better performance when we tune $\phi \cup \{\varphi_t\}$ in addition to $\alpha$ so that our default setting is tuning both of $\varphi$ and $\alpha$. Details in Sec.~\ref{sec:experiments}.}:
\begin{align}
    \varphi^* \gets  \mathop{\arg\min}\limits_{\Bar{\varphi}} L(\tau_t; \theta_0, \Bar{\varphi}).
\end{align}
There will be no further inference latency introduced by $\pi$-Tuning as the dimension of $\varphi^*$ is as same as before.

\begin{table*}[t]
\renewcommand\arraystretch{1.5}
\center
\small
\vskip 0.1in
\begin{adjustbox}{max width=1.\textwidth}
\begin{tabular}{lccccccccc}
\toprule[1.3pt]
  Method
  & Food101
  & Caltech101
  & DTD
  & EuroSAT
  & Aircraft
  & Flowers102
  & Pets
  & Cars
  & AVG
 
  \\
\Xhline{0.5pt}
\multicolumn{10}{l}{\textit{Multimodal Pretrained Baseline Models}} \\
    CLIP~\cite{radford2021learning}
    & 85.49
    & 93.76
    & 73.40
    & 95.70
    & 40.02
    & 94.94
    & 79.61
    & 62.84
    & 78.22
    \\
    FLAVA~\cite{amanpreet2022FLAVA}
    & 88.51
    & 95.74
    & 77.29
    & 97.26
    & 47.31
    & 96.37
    & 84.82
    & 70.87
    & 82.27
    \\
\Xhline{0.5pt}
\multicolumn{10}{l}{\textit{16-shot on OFA-Base}} \\
    Adapter
    & 69.27
    & 92.33
    & 54.31
    & 31.49
    & 31.32
    & 93.06
    & 77.81
    & 42.10
    & 61.46
    \\
    \baseline{$\pi$-Adapter}
    & \baseline{69.85\textred{(+0.58)}}
    & \baseline{92.74\textred{(+0.41)}}
    & \baseline{57.33\textred{(+3.02)}}
    & \baseline{36.49\textred{(+5.00)}}
    & \baseline{43.71\textred{(+16.39)}}
    & \baseline{94.52\textred{(+1.46)}}
    & \baseline{79.39\textred{(+1.58)}}
    & \baseline{54.04\textred{(+11.94)}}
    & \baseline{66.01\textred{(+4.55)}}
    \\
\Xhline{0.5pt}
\multicolumn{10}{l}{\textit{full data on OFA-Base}} \\
    Adapter
    & 85.77
    & 95.17
    & 72.75
    & 93.01
    & 45.24
    & 97.52
    & 89.26
    & 53.71
    & 79.05
    \\
    \baseline{$\pi$-Adapter}
    & \baseline{86.16\textred{(+0.39)}}
    & \baseline{95.82\textred{(+0.65)}}
    & \baseline{73.70\textred{(+0.95)}}
    & \baseline{93.94\textred{(+0.93)}}
    & \baseline{52.42\textred{(+7.18)}}
    & \baseline{98.05\textred{(+0.53)}}
    & \baseline{90.35\textred{(+1.09)}}
    & \baseline{61.30\textred{(+7.59)}}
    & \baseline{81.47\textred{(+2.42)}}
    \\
  
\bottomrule[1.3pt]
\end{tabular}
\end{adjustbox}
\caption{Experimental results on eight common vision tasks. We evaluate $\pi$-Tuning in both few-shot and full-data scenarios. The predicted auxiliary experts for vision tasks all contain image captioning, verifying the cross-modal transfer benefits.}
\label{tb:image}
\vskip -0.1in
\end{table*}
\begin{table*}[t]
\renewcommand\arraystretch{1.2}
\center
\small
\vskip 0.1in
\begin{adjustbox}{max width=1.\textwidth}
\begin{tabular}{lccccccc}
\toprule[1.3pt]
  Method
  & MNLI
  & QQP
  & MRPC
  & QNLI
  & RTE
  & SST2
  & AVG
 
  \\

\Xhline{0.5pt}
\multicolumn{8}{l}{\textit{Multimodal Pretrained Baseline Models}} \\
    VisualBERT~\cite{Liunian2019VisualBERT}
    & 81.6
    & 89.4
    & 71.9
    & 87.0
    & 56.6
    & 89.4
    & 79.3
    \\
    UNITER~\cite{yen2020UNITER}
    & 80.9
    & 89.2
    & 69.3
    & 86.0
    & 55.6
    & 89.7
    & 78.5
    \\
    Uni-Perceiver~\cite{Xizhou2022Uni-Perceiver}
    & 81.7
    & 87.1
    & 86.6
    & 89.9
    & 64.3
    & 90.2
    & 83.3
    \\

\Xhline{0.5pt}
\multicolumn{8}{l}{\textit{zero shot on OFA-Large}} \\
    OFA-L
    & 37.12
    & 37.31
    & 62.99
    & 49.50
    & 49.46
    & 55.85
    & 36.53
    \\
    \baseline{$\pi$-Adapter}
    & \baseline{44.36\textred{(+7.24)}}
    & \baseline{61.68\textred{(+24.37)}}
    & \baseline{69.85\textred{(+6.86)}}
    & \baseline{55.98\textred{(+6.48)}}
    & \baseline{51.99\textred{(+2.53)}}
    & \baseline{56.77\textred{(+0.92)}}
    & \baseline{42.58\textred{(+6.05)}}
    \\

\Xhline{0.5pt}
\multicolumn{8}{l}{\textit{full data on OFA-Large}} \\
    Adapter
    & 85.99
    & 90.70
    & 87.50
    & 92.49
    & 72.20
    & 93.81
    & 87.12
    \\
    \baseline{$\pi$-Adapter}
    & \baseline{86.06\textred{(+0.07)}}
    & \baseline{91.16\textred{(+0.46)}}
    & \baseline{87.75\textred{(+0.25)}}
    & \baseline{92.66\textred{(+0.17)}}
    & \baseline{76.53\textred{(+4.33)}}
    & \baseline{93.81\textcolor{gray}{(+0.00)}}
    & \baseline{88.00\textred{(+0.88)}}
    \\
\Xhline{0.5pt}
\multicolumn{8}{l}{\textit{full data on T5-Base}} \\
    Adapter
    & 86.03
    & 91.02
    & 89.71
    & 92.51
    & 73.57
    & 94.72
    & 87.93
    \\
    \baseline{$\pi$-Adapter}
    & \baseline{86.19\textred{(+0.16)}}
    & \baseline{91.13\textred{(+0.11)}}
    & \baseline{90.20\textred{(+0.49)}}
    & \baseline{92.62\textred{(+0.11)}}
    & \baseline{82.86\textred{(+9.29)}}
    & \baseline{95.07\textred{(+0.35)}}
    & \baseline{89.68\textred{(+1.75)}}
    \\
  
\bottomrule[1.3pt]
\end{tabular}
\end{adjustbox}
\caption{Experimental results on natural language understanding tasks from the GLUE benchmark~\cite{wang2018glue}. We experiment with $\pi$-Tuning in both zero-shot and full-data settings with respect to two backbone models, OFA and T5. }
\label{tb:language}
\vskip -0.1in
\end{table*}

\paragraph{Relationship between FIM and landscape.}
\label{sec:3.3}
We will provide complementary analytical analysis into the effectiveness of $\pi$-Tuning to show that tasks with similar FIM lead to close local optimums so that the similarity of FIM can be a nice indicator for the effectiveness of interpolation. For simplicity, we consider two tasks $\tau_1$ and $\tau_2$ with loss function of cross-entropy, which is formulated as $L=\mathop{\mathbb{E}}\limits_{(x,y) \sim P}[-\log P_\varphi(y|x)]$. Let $\varphi_1$ and $\varphi_2$ be the corresponding local minimum for $\tau_1$ and $\tau_2$, optimized from the same initial parameters $\varphi_0$, respectively. The FIM of each task can be seen as the negative Hessian matrix of loss, which serves as the second derivative of the loss function. 

We now state our main theorem in Theorem \ref{th1}, demonstrating the distance between $\varphi_1$ and $\varphi_2$ can be bounded when they have similar non-singular FIM along the linear path and their gradient at $\varphi_0$ is close. The detailed derivation is presented in Appendix \ref{sec:proof}.
\begin{theorem}\label{th1}
  Assume that the gradient of two tasks at the initial parameters $\varphi_0$ is close and we have similar non-singular FIM of two tasks along the linear path.
  Then the gap between two local minimal can be controlled by a constant $C$, \textit{i.e.}, 
  $||\varphi_1-\varphi_2||_2 \le C. $
\end{theorem}
We note that, as previous works revealed that fine-tuned checkpoints initialized from the same pre-train model lie in the same basin of the error landscape~\cite{neyshabur2020being}, we further found that PETL methods trained for similar tasks also have this property. Our method extends weight averaging on fine-tuned checkpoints under varied hyperparameter configurations for a single task~\cite{wortsman2022model} to various PETL methods and various tasks. As combined from different domains, our method shows a more robust performance compared to fine-tuning and original PETL methods when distribution shifts.

\label{sec:method}

\section{Experiments}
This section presents our key experimental findings. We begin with experimental settings (described in Sec.~\ref{sec:experimental_setting}), and then verify the effectiveness of $\pi$-Tuning on both cross-modal and uni-modal tasks (described in Sec.~\ref{sec:pi_tuning}). Next, we give the rationality of the similarity measurement and the interpolation operation (described in Sec.~\ref{sec:task_relation}). Afterward, we study the task-level transferability of the model after $\pi$-tuning(described in Sec.~\ref{sec:cross_task}). Finally, ablation studies of the key design choices are presented (described in Sec.~\ref{sec:ablation}).



\subsection{Experimental Settings}
\paragraph{Implementation details.} We choose the vision-language foundation model OFA \cite{wang2022ofa} as the pretrained model, with the mostly-used base-size and large-size version, whose parameters are 180M and 470M, respectively. We set the default number of auxiliary experts $k=2$. We follow the few-shot data split used in ~\citet{zhou2022learning}. More experimental details are provided in Appendix \ref{sec:appendix_experimental_setting}. 

\paragraph{Multimodal task pool.} We evaluate $\pi$-Tuning on a task pool across vision, language, and VL tasks. For vision tasks, we choose 8 common vision recognition tasks. For language tasks, we evaluate the model on 8 tasks from GLUE~\cite{wang2018glue}. For VL tasks, we experiment on both understanding and generation tasks, including RefCOCO, RefCOCO+~\cite{yu2016modeling}, RefCOCOg~\cite{mao2016generation}, VQAv2~\cite{goyal2017making}, SNLI-VE~\cite{xie2019visual} and COCO image captioning~\cite{chen2015microsoft}. We follow the evaluation metrics in~\citet{wang2022ofa} for VL and language tasks and~\citet{yang2022prompt} for vision tasks. 

\paragraph{Compared methods.} We compare $\pi$-tuning with finetuning and other four types of state-of-the-art PETL methods: 

\begin{itemize}
    \item Bitfit~\cite{zaken2021bitfit} optimizes bias terms in all linear layers at every Transformer layer.
    
    \item Adapter tuning~\cite{houlsby2019parameter} insert adapters between transformer layers, consists of a down-project $W_{\rm down} \in \mathbb{R}^{h \times r}$, followed by a nonlinear activation function, and an up-project $W_{\rm up} \in \mathbb{R}^{r \times h}$, where $h$ is the hidden size of the transformer model and $r$ is a hyperparameter of adapters as bottleneck dimension. We set $r=128$ in all experiments. 
    
    \item Prompt tuning~\cite{li2021prefix} prepends vectors to keys and values of the attention module at every layer. Specifically, there is a tunable prompt vector embedding $P\in \mathbb{R}^{L\times l \times h}$, where $L$ is the number of layers and $l$ is a hyper-parameter to define prefix vector length, to retrieve each prefix vector at every layer. We follow the setting of $l$ in ~\citet{yang2022prompt}.
    
    \item LoRA~\cite{hu2021lora} decomposes weight matrices of transformer layers as trainable parameters to approximate the weight updates. For a linear projection $h=Wx$ where $W\in \mathbb{R}^{d\times s}$, LoRA adds two trainable matrices  $B\in \mathbb{R}^{d\times r}$ and $A\in \mathbb{R}^{r\times s}$, where the rank is a hyperparameter and $r \ll \min (d, s)$ so that the linear projection is modified as $h=Wx + BAx$. We set $r=16$ in all experiments.
\end{itemize}

We refer $\pi$-tuning applied on adapters to $\pi$-Adapter, similarly for $\pi$-Prompt and $\pi$-LoRA.  We primarily use $\pi$-Adapter for experiments since its superior performance.

\label{sec:experimental_setting}

\subsection{Comparison with PETL Methods}
\label{sec:pi_tuning}

\begin{figure*}[!t]
\vskip -0.1in
\centering
\subfigure[linear mode connectivity analysis on RefCOCOg.]{
    \includegraphics[width=0.65\columnwidth]{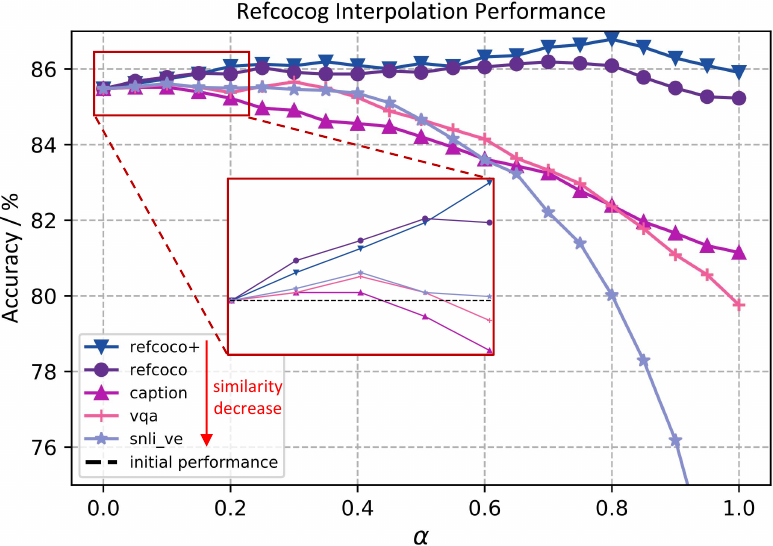}
    \label{line_graph}
}
\ \ 
\subfigure[relationship between similarity rank and interpolation accuracy.]{
    \includegraphics[width=0.65\columnwidth]{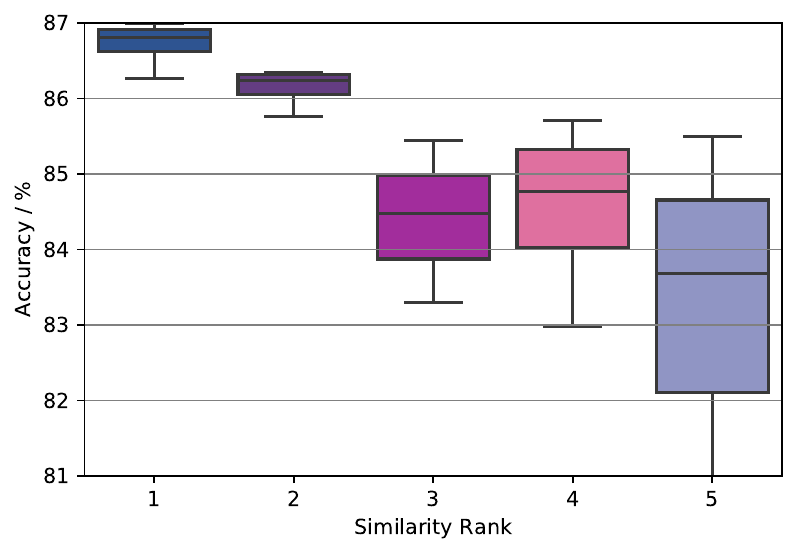}
    \label{FIM_performance_relation}
}
\subfigure[RefCOCOg test error landscape.]{
    {
    \includegraphics[width=0.62\columnwidth]{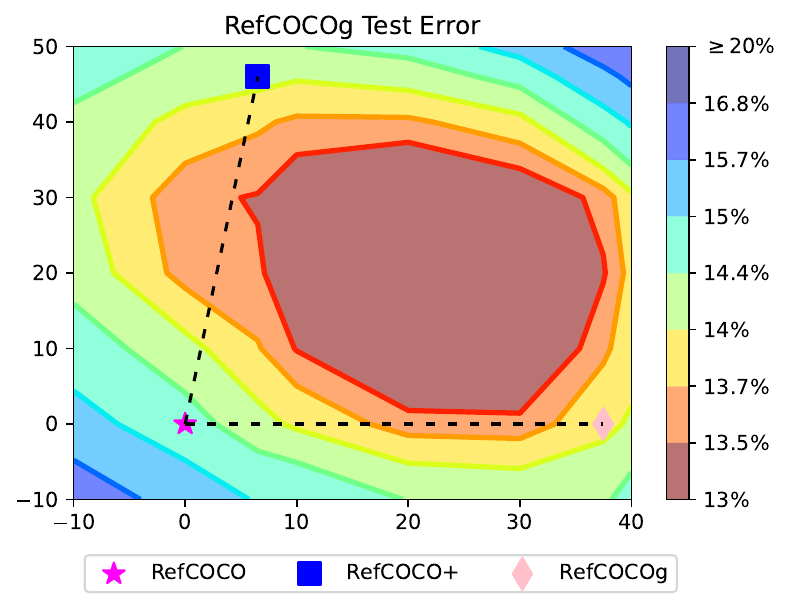}
    \label{fig:loss_landscape}
    }
}
\caption{(a) Accuracy when linearly interpolating between the parameter-efficient tuning checkpoint of RefCOCOg and checkpoints of other tasks (RefCOCO+ and RefCOCO). The advantage of interpolating is correlated with the FIM similarity between tasks. The interpolation of similar tasks' performance maximum is higher than dissimilar tasks and needs a greater interpolation weight. (b) Direct transfer performance (accuracy at $\alpha=1$) also correlates to FIM similarity. (c) The test error surface on RefCOCOg~\cite{mao2016generation}, as a function of model weights in a two-dimensional subspace\protect\footnotemark. It shows that checkpoints of similar tasks lie in the same basin of the surface. Interpolation of those checkpoints could achieve lower test error on the surface. The visualization follows~\citet{garipov2018loss}, which derives an orthonormal basis $\hat{u},\hat{v}$ by three checkpoints.}

\label{refcocog_interpolation}
\vskip -0.1in

\end{figure*}

\begin{table}[t]
\renewcommand\arraystretch{1.4}
\center
\vskip 0.15in
\begin{adjustbox}{max width=1.\columnwidth}
\begin{tabular}{l|cccc}
\toprule[1.3pt]
Method & \makecell[c]{Training Time \\ (GPU hours)} & \makecell[c]{Throughput \\ (samples/sec)} & \makecell[c]{Deployment \\ Params (\%)} & \makecell[c]{Training \\ Params (\%)} 
\\
\Xhline{0.5pt}
    Bitfit & 203 & 19.04 & 0.04\% & 0.04\%
    \\
    LoRA & 215 & 17.78 & 0.99\% & 0.99\%
    \\
    Prompt Tuning & 292 & 18.45 & 1.03\% & 1.03\%
    \\
    Adapter & 344 & 18.60 & 2.60\% & 2.60\%
    \\
    \baseline{$\pi$-Adapter} & \baseline{17} & \baseline{18.60} & \baseline{2.60\%} & \baseline{7.42\%}
    \\
  
\bottomrule[1.3pt]
\end{tabular}
\end{adjustbox}
\caption{Computational costs of different methods on RefCOCO, including the parameter proportion for tunable and deployment parts of the network, as well as wall time during training versus inference. Training time is measured by A100 GPU hours. The inference cost is indicated by the throughput, which is the samples processed per second by a single A100 GPU.}
\vskip -0.1in
\label{tb:computation_cost}
\end{table}

\paragraph{Multi-/uni-modal tasks.}
Table~\ref{tb:multimodal_results} shows the results on 6 multimodal datasets of $\pi$-tuning as well as previous SOTA results. We can see that (1) $\pi$-Tuning improves original adapters across all tasks consistently; (2) $\pi$-Tuning outperforms all the other parameter-efficient tuning methods across all tasks, without introducing additional inference latency, indicating that $\pi$-Tuning is stronger in storage-constrained scenarios; (3) $\pi$-Tuning achieves better or comparable results towards fully finetuning on 5 out of 6 datasets for the large-size model, while using significantly fewer trainable parameters.
Although inferior to finetuning for base size, $\pi$-Tuning still offers non-trivial performance gain for PETL methods, similar to the results in~\citet{yang2022prompt}.

We further show $\pi$-Tuning is competitive on uni-modal tasks, including both vision and language tasks. The outcome is demonstrated in Table \ref{tb:image} and Table \ref{tb:language}. $\pi$-Tuning achieves a consistent performance gain. Specifically, we observe cross-modal transfer benefits in these experiments, where experts for image captioning can help with vision tasks and language entailment tasks can be benefited from experts for visual entailment. We attribute this to the similar semantics of tasks, as captioning can be regarded as a more detailed classification task and some visual entailment samples may directly rely on language entailment.

We list and discuss the computational costs of different PETL methods during training versus inference. 
As demonstrated in Table \ref{tb:computation_cost}, our $\pi$-Adapter shows the same throughput and parameters during inference as original adapters. $\pi$-Adapter requires more parameters for training due to the interpolation of multiple experts from similar tasks, which is the core insight of our paper. However, the training time for such an interpolation is quite short as the experts have been well-trained. We would like to clarify that the performance improvement of $\pi$-Adapter does not come from increasing tunable parameters. As shown in Table \ref{tb:ablation} of our paper, the performance decreases when solely scaling up the tunable parameters without initialization from experts of similar tasks, indicating the effectiveness of transferring knowledge from similar tasks’ experts.

\begin{table}[t]
\renewcommand\arraystretch{1.4}
\setlength\tabcolsep{3.2pt}
\center
\vskip 0.15in
\begin{adjustbox}{max width=1.\columnwidth}
\begin{tabular}{l|ccc|ccc|cc}
\toprule[1.3pt]
  \multirow{2}*{Method}
  &\multicolumn{3}{c|}{RefCOCO}
  &\multicolumn{3}{c|}{RefCOCO+}
&\multicolumn{2}{c}{RefCOCOg}
 
  \\
  & val & testA & testB
  & val & testA & testB
  & val-u & test-u
    \\
\Xhline{0.5pt}
    Prompt Tuning
    & 84.53 & 85.21 & 77.36
    & 76.34 & 81.44 & 67.68
    & 75.61 & 76.57

    \\
    \baseline{$\pi$-Prompt}
    & \textbf{\baseline{85.75}} & \textbf{\baseline{88.85}} & \textbf{\baseline{79.67}}
    & \textbf{\baseline{77.84}} & \textbf{\baseline{83.09}} & \textbf{\baseline{69.61}}
    & \textbf{\baseline{77.41}} & \textbf{\baseline{78.06}} 
    \\
    \hline
    LoRA
    & 81.91 & 85.89 & 76.90
    & 72.29 & 79.22 & 62.28
    & 72.55 & 73.26 

    \\
    \baseline{$\pi$-LoRA}
    & \textbf{\baseline{85.82}} & \textbf{\baseline{88.97}} & \textbf{\baseline{81.77}}
    & \textbf{\baseline{78.41}} & \textbf{\baseline{83.95}} & \textbf{\baseline{69.22}}
    & \textbf{\baseline{77.53}} & \textbf{\baseline{78.38}} 

    \\
\bottomrule[1.3pt]
\end{tabular}
\end{adjustbox}
\caption{$\pi$-Tuning results based on Prompt Tuning and LoRA for the base size model on referring expression comprehension datasets, indicating that $\pi$-Tuning is agnostic to PETL methods and can provide a consistent gain.}
\label{tb:extend}
\vskip -0.2in
\end{table}
\paragraph{Few/zero-shot setting.}
To have a better understanding of the benefit of knowledge transferred from similar task experts, we experiment under two low-data settings, zero-shot natural language understanding and few-shot image classification to validate the effectiveness of $\pi$-Tuning when data is scarce. We show the 16-shot image classification results in Table~\ref{tb:image}, where $\pi$-Tuning brings more improvement compared to the full-data setting, with an average of 4.55\% accuracy improvement across 8 tasks. For the zero-shot setting, as we can not derive the target expert, 
we test the zeros-shot transfer performance of the target task using the expert of its nearest neighbor (most similar task), 
The zero-shot natural language understanding results are presented in Table~\ref{tb:language}, where the average improvement across tasks is more than that under the full-data setting.

\paragraph{Pretrained foundation models.}
We also extend $\pi$-Tuning on a uni-modal pretrained foundation model, T5~\cite{raffel2020exploring}, to solve natural language understanding tasks. We demonstrate the results in Table \ref{tb:language}, where we find that RTE is the task that benefits most with an improvement of 9.29\% accuracy.

\paragraph{Experts from PETL methods.} 
In Table~\ref{tb:extend}, experiments on RefCOCO, RefCOCO+, and RefCOCOg datasets with the base-size model show that $\pi$-Tuning is agnostic to PETL methods, providing significant and consistent performance gain for LoRA and Prompt Tuning. $\pi$-LoRA improves a large margin of $4.20\%$ on average accuracy across three datasets towards original LoRA and $1.94\%$ for $\pi$-Prompt towards original Prompt Tuning.


\subsection{Task Relationships}
\label{sec:task_relation}

\paragraph{Linear mode connectivity.}
As defined in~\citet{frankle2020linear}, two parameters $w_1$ and $w_2$ are linear mode connectivity if the error barrier \cite{garipov2018loss,draxler2018essentially} height $\approx 0$ along the linear path between them.
To provide our intuition, we first study the performance along the linear path between the target task expert $\varphi_t$ and experts for other tasks from the task set to see whether they have linear mode connectivity. We use adapters as parameter-efficient experts to store task information and RegCOCOg as the target task. We vary interpolation coefficient $\alpha$ from 0 to 1, with an interval of 0.05, and the combined expert $\Bar{\varphi}$ is given by $(1-\alpha)\varphi_t+\alpha\varphi$, $\varphi \in \Phi$, which is a special case of $k=1$ as we defined in Sec.~\ref{sec:3.2}.

As shown in Fig.~\ref{line_graph}, the model can benefit from interpolation compared to the original target task expert $\varphi_t$ when $\alpha=0$ and there is linear mode connectivity between the target task expert and experts of similar tasks, while for dissimilar tasks there is large error barrier along the linear path. Furthermore, these results suggest that (1) experts with higher similarity need a greater interpolation coefficient, and (2) direct transfer~($\alpha=1$) performance and maximum interpolation performance correlate with similarity.

\paragraph{Task similarity measurement.}
To investigate the correlation between task similarity and interpolation performance, we consider interpolation performance on the validation and test split of RefCOCO, RefCOCO+, and RefCOCOg. We compare the average accuracy across those splits for the interpolated experts with different similarities. The results are illustrated in Fig.~\ref{FIM_performance_relation}, where the interpolation accuracy increases when experts with higher similarity to the target task are interpolated. We also visualize the task similarity of each task pair from our task space, which consists of both unimodal and multimodal tasks, in Fig. \ref{task_heatmap}. The results illustrate that task embedding of vision tasks is closer to vision\&language tasks compared to language and the embedding of image captioning task is similar to almost all the vision and VL tasks. We assume that is one of the reasons why image captioning is important in image-text multimodal pretraining.

\paragraph{Error landscape visualization.} We also visualize two slices of the test error landscape when interpolating experts in Fig. \ref{fig:loss_landscape}, where the error contours are basin-shaped and none of the individual domain-specific experts is optimal. The results suggest that interpolation may reach a better point in the basin of the landscape. 
\begin{figure}[!t]
\begin{center}
\centerline{\includegraphics[width=0.95\columnwidth]{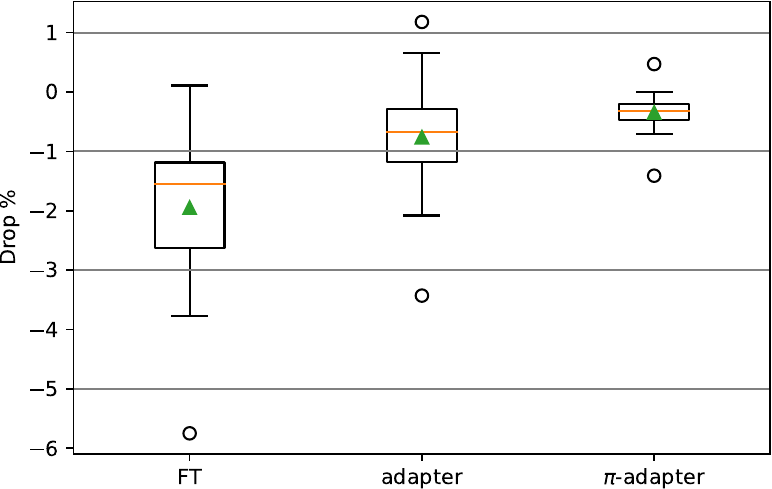}}
\caption{Relative performance drop (\%) of $\pi$-Adapter compared to domain-specific models on referring expression comprehension datasets (RefCOCO, RefCOCO+, RefCOCOg). FT represents finetuning and the green triangles denote the mean.}
\label{boxplot}
\end{center}
\vskip -0.15in
\end{figure}

\subsection{Cross-Task Transferability}
\label{sec:cross_task}
\textbf{Distribution shift.}
The boxplot in Fig.~\ref{boxplot} summarizes the relative performance drop of $\pi$-Adapter due to test distribution shift compared to domain-specific models, which are optimized directly on the test domain. Specifically, we conduct experiments on referring expression comprehension datasets. As combined with experts trained from other domains, $\pi$-Adapter shows more robust performance than adapters and fine-tuning trained on the specific domain when the test distribution is shifted. Even though $\pi$-Adapter is further tuned on another specific domain, it still can be benefited from the interpolation with experts trained from other domains, while direct optimization of adapters or fine-tuning all the parameters may fit on the specific domain, lacking domain generalization ability.

\begin{table}[t]
\renewcommand\arraystretch{1.4}
\center
\begin{adjustbox}{max width=1.\columnwidth}
\begin{tabular}{l|ccc|ccc|cc}
\toprule[1.3pt]
\multirow{2}*{Method}
  &\multicolumn{3}{c|}{RefCOCO}
  &\multicolumn{3}{c|}{RefCOCO+}
  &\multicolumn{2}{c}{RefCOCOg}
 
  \\
  & val & testA & testB
  & val & testA & testB
  & val-u & test-u
    \\
\Xhline{0.5pt}
    \multicolumn{9}{l}{\textit{OFA-Base}} 
    \\
    Adapter
    & {86.34} & {89.61} & {80.82}
    & {74.60} & {83.20} & {69.79}
    & {79.74} & {80.65}
    \\
    \baseline{$\pi$-Adapter}
    & \baseline{\textbf{87.12}} & \baseline{\textbf{90.30}} & \baseline{\textbf{82.16}}
    & \baseline{\textbf{79.46}} & \baseline{\textbf{84.63}} & \baseline{\textbf{71.43}}
    & \baseline{\textbf{80.84}} & \baseline{\textbf{82.00}}
    \\
\Xhline{0.5pt}
    \multicolumn{9}{l}{\textit{OFA-Large}} 
    \\
    Adapter
    & {90.00} & {92.88} & {85.24}
    & {83.81} & {89.08} & {76.54}
    & {85.99} & {85.96}
    \\
    \baseline{$\pi$-Adapter}
    & \baseline{\textbf{{90.55}}} & \baseline{\textbf{{93.12}}} & \baseline{\textbf{{85.85}}}
    & \baseline{\textbf{84.77}} & \baseline{\textbf{{90.31}}} & \baseline{\textbf{77.68}}
    & \baseline{{\textbf{86.91}}} & \baseline{{\textbf{86.88}}}

    \\
  
\bottomrule[1.3pt]
\end{tabular}
\end{adjustbox}
\caption{$\pi$-Tuning can improve the performance compared to directly optimization in a multitask setting, achieving task specific model's performance across all task splits.}
\label{tb:multi_task}
\end{table}

\textbf{Multi-task learning.} As interpolation with experts of different tasks can be regarded as an implicit multitask learning, we experiment with $\pi$-Tuning in a multitask setting, demonstrated in Table \ref{tb:multi_task} shows. Compared to direct multitask learning on different tasks, we observe that $\pi$-Tuning performs better in all task splits, even exceeding the task-specific model's performance. 

\subsection{Ablation Study}

\begin{table}[t]
\renewcommand\arraystretch{1.4}
\center
\vskip 0.15in
\begin{adjustbox}{max width=1.\columnwidth}
\begin{tabular}{l|ccc|ccc|cc}
\toprule[1.3pt]
\multirow{2}*{Method}
  &\multicolumn{3}{c|}{RefCOCO}
  &\multicolumn{3}{c|}{RefCOCO+}
  &\multicolumn{2}{c}{RefCOCOg}
 
  \\
  & val & testA & testB
  & val & testA & testB
  & val-u & test-u
    \\
\Xhline{0.5pt}
    w/o init.
    & {89.95} & {92.36} & {84.79}
    & {83.81} & {89.31} & {76.87}
    & {85.36} & {85.68}
    \\
    only scale 
    & \textbf{90.67} & {92.75} & {85.55}
    & {84.64} & {89.71} & {77.03}
    & {86.34} & {86.75}
    \\
    
    \baseline{$\pi$-Adapter}
    & \baseline{{{90.49}}} & \baseline{\textbf{{92.93}}} & \baseline{\textbf{{85.91}}}
    & \baseline{\textbf{84.92}} & \baseline{\textbf{{90.03}}} & \baseline{\textbf{77.91}}
    & \baseline{{\textbf{86.60}}} & \baseline{{\textbf{86.92}}}

    \\
  
\bottomrule[1.3pt]
\end{tabular}
\end{adjustbox}
\caption{Ablation of $\pi$-Tuning. ``w/o init.” denotes we randomly initialize auxiliary experts and ``only scale” means we only tune the interpolation weight}
\label{tb:ablation}
\end{table}
\begin{figure}[!t]
\vskip -0.1in
\begin{center}
\centerline{\includegraphics[width=0.95\columnwidth]{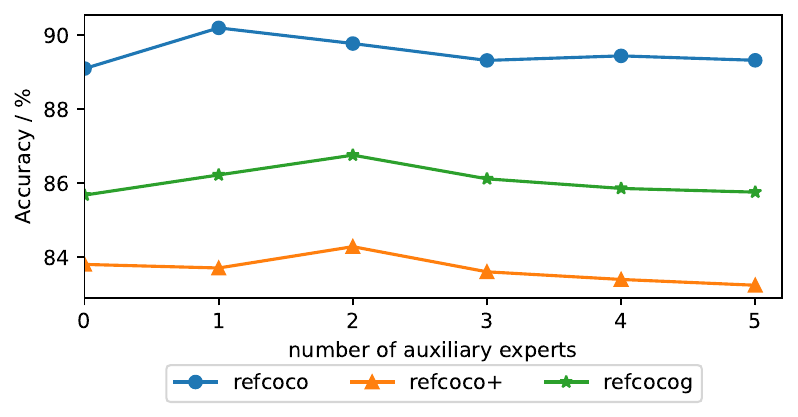}}
\vskip -0.1in
\caption{The ablation of the number of experts used in $\pi$-Tuning. The performance first rises to a maximum when the number of auxiliary experts increases and then gradually falls.}
\label{k_number_ablation}
\end{center}
\vskip -0.2in
\end{figure}

\label{sec:ablation}
\paragraph{The ablation of $\pi$-Tuning. } The ablation study of the proposed method is shown in Table \ref{tb:ablation}. ``w/o init." denotes we randomly initialize auxiliary experts, which performs even worse than the original adapter. ``Only scale" means we only tune the interpolation weight $\alpha$, achieving better results than the original adapter while slightly worse than $\pi$-Adapter, which is reasonable as it only updates a subspace of parameters in $\pi$-Tuning. 

\paragraph{The ablation of the number of experts. } We experimentally explored the relationship between performance on 3 target tasks and the number of auxiliary experts for each of them, demonstrated in Fig. \ref{k_number_ablation}. We observe that the performance first rises to a maximum when $k$ increases and then gradually falls, and the best value of k is around 2. We conjecture that the reason may lie in that as interpolation is conducted in descending order of task similarity, early interpolation with highly similar tasks can improve performance while following mixing dissimilar experts has negative inference on the target task.

\label{sec:experiments}

\section{Scope and Limitations}

While this work has so far demonstrated that interpolation of the experts learned from similar tasks is a useful technique for improving accuracy and robustness, this section explores the limitations of the approach which are expected to be addressed in future studies. 

\textbf{Task similarity measurement.} The task similarities in $\pi$-Tuning are measured by a diagonal approximation of FIM following the practice in ~\cite{achille2019task2vec}. While it works well for multi-task interpolation in our paper, there can be other solutions without the unmanageable computational overhead of a full FIM. Recently, ~\citet{vu2022spot, zhou2022efficiently} demonstrate that the parameters of the expert itself can be regarded as the task embedding, which can be a substitute solution to be applied to our approach.

\textbf{Applicability.} We verify the effectiveness of $\pi$-Tuning on OFA-Base/Large and T5-Base foundation models due to the limitation of computational resources.  We notice that parameter-efficient experts work well in the field of AIGC, like ~\citet{mou2023t2i}, which combines several adapters tuned for different conditions to achieve controllable image synthesis. More evaluations on much larger backbones and more fields would make the work more practical. And the number of tasks used for interpolation is manually designated in the current version. It is promising to determine the number of tasks adaptively.
\label{sec:limitation}

\section{Related Work}
\noindent\textbf{Parameter efficient transfer learning.}
In recent years, large-scale models pre-training on huge datasets have shown great capability in downstream tasks.
However, the cost of directly finetuning the large-scale models is expensive in memory and storage.
To mitigate this problem, researchers proposed several PETL methods to adapt large-scale pre-trained models with a few trainable parameters.
~\citet{houlsby2019parameter} proposed inserting adapter layers between transformer layers.
~\citet{hu2021lora} proposed injecting trainable low-rank matrices into transformer layers to approximate the weight updates.
~\citet{li2021prefix,lester2021the,liu2021ptuning} proposed optimizing the input word embeddings.
However, these PETL methods tend to be limited to the target downstream task, ignoring the potential complementarity between different tasks and modalities.

\noindent\textbf{Identifying beneficial task relationships.}
~\citet{poth2021pre,armen2021muppet, lu202012} have shown that pre-finetuning, an additional large-scale multitask learning stage between pre-training and finetuning, can significantly improve performance on the downstream tasks.
However, 
multitask learning may appear conflicts between different tasks, or even have an adverse impact on target tasks.
To identify beneficial task relationships,~\citet{joachim2017identify} rely on features derived from learning curves and~\citet{hector2017when} proposed using characteristics of datasets.
~\citet{amir2018taskonomy} proposed representing the CV task relationships as an affinity matrix and optimizing the affinity matrix to get the most training-efficient task set. ~\citet{achille2019task2vec,vu2020exploring} proposed using Fisher information matrix to extract task relationships.

\noindent\textbf{Averaging model weights.}
Ensembling the outputs of multiple models is an important method for improving the performance of deep learning models~\cite{thomas2000ensemble, balaji2017simple}.
However, when the size of the model is huge, ensembling the outputs of multiple models would be prohibitively expensive.
Unlike model ensembles, averaging the weights of models is a foundational technique in convex optimization and deep learning without any extra costs at inference time.
~\citet{neyshabur2020being} find that when two models are fine-tuned from the same pre-trained initialization, the interpolated model would get at least the accuracy of the endpoints.
~\citet{wortsman2022model} proposed to produce a better model by averaging the weights of models finetuned with different hyperparameter configurations.
\label{sec:related}
\section{Conclusion}
This paper presents $\pi$-Tuning, a new parameter-efficient transfer learning framework, to exploit universal representation across modalities by taking benefits from similar tasks.
Given a target task, $\pi$-tuning first retrieves several similar tasks in arbitrary modalities and obtains a more transferable model by the interpolation of cross-modal experts. We theoretically and experimentally demonstrate the retrieved tasks tend to locate around the same basin on the error landscape. Experiments on diverse unimodal and multimodal tasks verify the cross-task transferability of the model in both full-data and low-data scenarios.    

\vspace{-.3em}
\paragraph{Acknowledgement.}
This paper is partially supported by the National Key R\&D Program of China No.2022ZD0161000 and the General Research Fund of Hong Kong No.17200622. We thank Lehan Wang for her technical assistance, and Zhuoning Guo, Wenhao Lin, and Kaiyu Huang for their helpful comments.





\label{sec:conclusion}
\bibliography{11_references}
\bibliographystyle{icml2023}
\newpage
\appendix
\onecolumn

\section{Experimentin Setups}
\label{sec:appendix_experimental_setting}
\subsection{Experimental Setting for VL Tasks}
\paragraph{Referring Expression Comprehension} We report the standard metric ACC@0.5 on the validation and test sets. Epochs are set to 100, dropout is set to 0.1, warmup rate is set to 0.06, and label smoothing rate is set to 0.1. For prompt tuning, we follow the experimental setting used in ~\citet{yang2022prompt}, where the batch size is set to 128, the learning rate is set to 0.03, and the prompt length is set to 100. For LoRA, the batch size is set to 256 for the base, and 1024 for the large, the learning rate is set to 1e-4, and the rank is set to 16. For adapters, the batch size is set to 1024, the learning rate is set to 1e-4, and the bottleneck dimension is set to 128. 
\paragraph{Visual Entailment} We report accuracy on both dev and test sets. Dropout is set to 0.1, warmup rate is set to 0.06, and label smoothing rate is set to 0.1. For prompt tuning, we follow the experimental settings in ~\citet{yang2022prompt}, where the batch size is set to 128, the learning rate is set to 0.03, epochs are set to 100 and the prompt length is set to 64. For LoRA, the batch size is set to 256 for the base, and 512 for the large, epochs are 100, the learning rate is set to 1e-4, and the rank is set to 16. For adapters, the batch size is set to 512, epochs are set to 10, the learning rate is set to 1e-4, and the bottleneck dimension is set to 128.
\paragraph{Image Captioning} We report BLEU@4, METEOR, CIDEr, and SPICE scores on the Karpathy test split.  Dropout is set to 0.1, warmup rate is set to 0.06, and label smoothing rate is set to 0.1. For prompt tuning, we follow the experimental setting used in ~\citet{yang2022prompt}, where the batch size is set to 256, the learning rate is set to 0.03, epochs are set to 100 and the prompt length is set to 64. For LoRA, the batch size is set to 128, epochs are 100, the learning rate is set to 1e-4, and the rank is set to 16. For adapters, the batch size is set to 512 for base and 128 for large, epochs are set to 10, the learning rate is set to 1e-4, and the bottleneck dimension is set to 128.
\paragraph{Visual Question Answering} We conduct experiments on VQA 2.0 and report the score on the test-dev and test-std set. Dropout is set to 0.1, warmup rate is set to 0.04, Exponential Moving Average(EMA) with a decay rate is set to 0.9999, and label smoothing rate is set to 0.1. For prompt tuning, we follow the experimental setting used in ~\citet{yang2022prompt}, where the batch size is set to 256, the learning rate is set to 0.03, epochs are set to 100 and the prompt length is set to 10. For LoRA, the batch size is set to 128, epochs are 100, the learning rate is set to 1e-4, and the rank is set to 16. For adapters, the batch size is set to 256, epochs are set to 30 for base and 15 for large, and the bottleneck dimension is set to 128. After finetuning, we use beam search to generate our answer. 

After we retrieve auxiliary experts, we apply the same optimization step as the initial experts to train the interpolation of experts. All of the experiments are conducted in A100 40G and V100 32G.

\subsection{Experimental Setting for Vision Tasks}
We select 8 common vision tasks from COOP~\cite{zhou2022learning}, following their splits. For few-shot image classification each adapter for 200 epochs, with a batch size of 64 and a learning rate of 1e-4. The ratio for label smoothing is 0.1. We follow the data augmentation used in ~\citet{wang2022ofa}, where the same random resize cropping, random flipping, RandAug, and random erasing transformations are conducted. We use Mixup and CutMix with an overall 0.5 probability to be performed for each batch and alpha is set to 0.8 and 1.0. 
\subsection{Experimental Setting for Language Tasks}
We select 6 language understanding tasks from GLUE benchmark~\cite{wang2018glue} as ~\citet{wang2022ofa}, including both single-sentence classification tasks and sentence-pair classification tasks. We reuse the instructions of each task used in ~\citet{wang2022ofa} in our experiments. For the hyper-parameters of adapters, we tune the training epochs among \{5, 7, 10\}, learning rate among \{0.03, 1e-4, 5e-5\}, batch size among \{32, 64, 128\}. We report the best performance on the development set for each task following ~\citet{wang2022ofa}.
\section{Analysis For Relationship Between FIM and Landscape}
\label{sec:proof}
Here we will give a detailed analysis of the relationship between FIM and local optimums. We begin this by restating and adding to the notation used in Sec. \ref{sec:3.1}.
\subsection{Notation and preliminaries
}
Let $\theta_0$ be the pre-trained weights and $\varphi_0$ be the initial extra added parameters. For a model with 
pre-trained weights $\theta_0$, extra added parameters $\varphi$ and input vector $x$, we let $P_\varphi(y|x)$ denote the model's output distribution. 
The two loss functions of tasks $\tau_i$ $(i=1,2)$ 
can be formulated as $L=\mathop{\mathbb{E}}\limits_{(x,y) \sim P_i}[-\log P_\varphi(y|x)]$, where $P_i$ is the corresponding empirical distribution of task $\tau_i$.
The FIM $F_\varphi^i$ of each task $\tau_i$ can be seen as the negative of the Hessian matrix of loss, which is formulated as $F_\varphi^i = -\nabla_\varphi^2 L^i_\varphi = \mathbb E_{x,y\sim P_i} [\nabla_\varphi \log P_\varphi(y|x) \nabla_\varphi \log P_\varphi(y|x)^T]$.
Let $\varphi_1$ and $\varphi_2$ be the corresponding local minimum for $\tau_1$ and $\tau_2$, optimized from the same initial parameters $\varphi_0$, respectively.

\subsection{Restatement of Assumption}
Before we start our proof of Theorem \ref{th1}, we introduce the assumptions mentioned in Section \ref{sec:3.3} formally.

\begin{assumption}\label{as2}
  The gradient of two loss functions are similar at the initial point ${\varphi_0}$, \textit{i.e.},
  there is a constant $C_1$ such that 
  $||\nabla_\varphi L_1(\varphi_0)- \nabla_\varphi L_2(\varphi_0)||_2 \le C_1. $ 
\end{assumption}
\begin{assumption}\label{as3}
  The Fisher information matrix in the linear line from $\varphi_0$ to $\varphi_1$ 
  for two tasks are similar,
  \textit{i.e.}, there is a constant $C_2$ for $\varphi$ along the line to have
  $||F^1_{\varphi}- F^2_{\varphi}||_2 \le C_2. $ 
\end{assumption}
\begin{assumption}\label{as4}
  The Fisher information matrix for task $\tau_2$ in the interpolation line is similar to 
   which 
  at $\varphi_1$, \textit{i.e.}, there is a constant $0<C_3<1$ for $\varphi$ along the interpolation line to have the following property
  $$||F^2_{\varphi}-F^2_{\varphi_1}||_1 \le \frac{1-C_3}{n_0||[F^2_{\varphi_1}]^{-1}||_2} = \frac{1-C_3}{n}\sqrt{\lambda_{\text{min}}([F^2_{\varphi_1}]^t F^2_{\varphi_1})}, 
  $$
  where $n_0$ is the dimension of extra added parameters, $\lambda_{\text{min}}$ means to get the smallest one of eigenvalues.
\end{assumption}

Assumption \ref{as2} and Assumption \ref{as3} are mild as Assumption \ref{as2} only requires the norm of gradient difference of two tasks in the initial parameters $\varphi_0$ can be bounded and \ref{as3} is the formal restatement of our requirement of two similar tasks.
Assumption \ref{as4} requires 
 FIM in the interpolation line varies little
for avoiding the Fisher information matrix degenerating since when $C_3 >1$ the FIM in the interpolation line can degenerate.
It corresponds to the non-singular requirement in Theorem \ref{th1}.
We add this condition for the reason that a quick change of FIM in the interpolation line will lead to difficulty in comparing 
FIM.

\subsection{Proof of Theorem \ref{th1}}
We now begin our proof of Theorem \ref{th1}:
\begin{proof}
Since $\varphi_1$ and $\varphi_2$ are local minimums for two loss functions, the gradient of loss there will be zero, \textit{i.e.}
$$ \nabla _\varphi L_i(\varphi_i) =0.  $$
Let $\varphi_t=\varphi_0+t(\varphi_1-\varphi_0)$ and 
$\hat\varphi_{t}=\varphi_1+t(\varphi_2-\varphi_1)$ for $0\le t\le 1$. $c=||[F^2_{\varphi_1}]^{-1}||_2$.

By the Newton-Leibniz formula, we have
\begin{flalign*}
&\ 0 =\nabla_\varphi L_1({\varphi_1}) - \nabla_\varphi L_2({\varphi_2}) \\
&\ = \nabla_\varphi L_1({\varphi_0}) - \int_0^1 F^1_{\varphi_t}({\varphi_1}-{\varphi_0})dt 
 - \nabla_\varphi L_2({\varphi_0}) + \int_0^1 F^2_{\varphi_t}({\varphi_1}-{\varphi_0})dt + \int_0^1 F^2_{\hat\varphi_{t}}({\varphi_2}-{\varphi_1})dt \\
&\ = \nabla_\varphi L_1({\varphi_0}) - \nabla_\varphi L_2({\varphi_0}) -  \int_0^1 [F^1_{\varphi_t}-F^2_{\varphi_t}]({\varphi_1}-{\varphi_0})dt + \int_0^1 F^2_{\hat\varphi_{t}}({\varphi_2}-{\varphi_1})dt.
\end{flalign*}

Moving the last term to the left, we will have the following statement
\begin{flalign*}
  &\ \left[\int_0^1 F^2_{\hat\varphi_{t}}dt\right]({\varphi_2}-{\varphi_1}) \\
  &\ = \nabla_\varphi L_2({\varphi_0}) - \nabla_\varphi L_1({\varphi_0}) +  \left[\int_0^1 [F^1_{\varphi_{t}}-F^2_{\varphi_{t}}]dt\right]({\varphi_1}-{\varphi_0}).
  \end{flalign*}

The formula on the left is 
the integration of FIM,
in the interpolation line,
multiplying the difference between two local minimums.
Thus we will check the difference between the integration and the FIM at $\varphi_1$.

  \begin{gather*}
  \left|\left|\int_0^1 F^2_{\hat\varphi_{t}}dt - F^2_{\varphi_1}\right|\right|_1=
  \left|\left|\int_0^1 F^2_{\hat\varphi_{t}} - F^2_{\varphi_1}dt\right|\right|_1 
  \le  \int_0^1 \left|\left|F^2_{\hat\varphi_{t}} - F^2_{\varphi_1}\right|\right|_1dt
  \le \int_0^1 \frac{1-{C_3}}{n_0||[F^2_{\varphi_1}]^{-1}||_2} dt = \frac{1-{C_3}}{n_0c}.
\end{gather*}

The last inequality rises from Assumption \ref{as4}.
  Let $H=\int_0^1 F^2_{\hat\varphi_{t}}dt$ , $H_0=F^2_{\varphi_1}$. 
  Then the statement above can be reformulated as $$\left|\left|H - H_0 \right|\right|_1\le \frac{1-{C_3}}{n_0||[F^2_{\varphi_1}]^{-1}||_2} = \frac{1-{C_3}}{n_0c}.$$
  Since $H_0 = F_{\varphi_1}^2$ is symmetric, $H_0$ can be diagonalized orthogonally. Thus we can assume $H_0=P\Lambda P^{-1}$, where $\Lambda$ is diagonal and $P$ is orthogonal.
We have$$\left|\left|P^{-1}HP - \Lambda \right|\right|_1 
  = \left|\left|P^{-1}(H - H_0)P \right|\right|_1
  \le \left|\left|P^{-1}\right|\right|_1 \cdot \left|\left|P\right|\right|_1 \cdot \left|\left|H - H_0\right|\right|_1;$$
  Notice that for any square matrix $M$ of size $n_0$, we have $||M||_1 \le \sqrt{n_0} ||M||_2$.
  We immediately obtain
\begin{flalign*}
\left|\left|P^{-1}\right|\right|_1 \cdot \left|\left|P\right|\right|_1 \cdot \left|\left|H - H_0\right|\right|_1  \le \sqrt{n_0}\left|\left|P^{-1}\right|\right|_2 \cdot \sqrt{n_0}\left|\left|P\right|\right|_2 \cdot \left|\left|H - H_0\right|\right|_1 \\
=n_0 \left|\left|H - H_0\right|\right|_1
\le \frac{1-{C_3}}{||H_0^{-1}||_2}= \frac{1-{C_3}}{||\Lambda^{-1}||_2}.
\end{flalign*}

The first equality rises from the properties of orthogonality of $P$ and $P^{-1}$.
According to Gershgorin's circle theorem, the eigenvalues of $P^{-1}HP$ can be covered by $n_0$
circles with centers of the diagonal elements of $\Lambda$ and 
radius of all $\left|\left|P^{-1}HP - \Lambda \right|\right|_1$. 
Since the diagonal elements of $\Lambda$ are of norms at least $\frac{1}{||\Lambda^{-1}||_2}$ and the radius is less than $\frac{1-C_3}{||\Lambda^{-1}||_2}$,
the eigenvalues of $P^{-1}HP$ are of 
norms at least $\frac{1}{||\Lambda^{-1}||_2}- \frac{1-{C_3}}{||\Lambda^{-1}||_2} = \frac{{C_3}}{||\Lambda^{-1}||_2}$.
Therefore we have $$||H^{-1}||_2 = ||(P^{-1}HP)^{-1}||_2 \le \frac{||\Lambda^{-1}||_2}{{C_3}}= \frac{||H_0^{-1}||_2}{{C_3}}=\frac{c}{C_3}.$$

Since we already have 
\begin{flalign*}
  &\ H({\varphi_2}-{\varphi_1}) = \nabla_\varphi L_2({\varphi_0}) - \nabla_\varphi L_1({\varphi_0}) +  \left[\int_0^1 [F^1_{\varphi_{t}}-F^2_{\varphi_{t}}]dt\right]({\varphi_1}-{\varphi_0}),
  \end{flalign*}

left multiplying the matrix $H^{-1}$ on both sides we will get the following statement
  \begin{flalign*}
    &\ {\varphi_2}-{\varphi_1} = H^{-1}\left[\nabla_\varphi L_2({\varphi_0}) - \nabla_\varphi L_1({\varphi_0}) +  \left[\int_0^1 [F^1_{\varphi_{t}}-F^2_{\varphi_{t}}]dt\right]({\varphi_1}-{\varphi_0})\right].
    \end{flalign*}
Thus the distance between two local minimums can be further bounded as
    \begin{gather}
       ||{\varphi_2}-{\varphi_1}||_2 = \left|\left|H^{-1}\left[\nabla_\varphi L_1({\varphi_0}) - \nabla_\varphi L_2({\varphi_0}) +  \left[\int_0^1 [F^1_{\varphi_{t}}-F^2_{\varphi_{t}}]dt\right]({\varphi_1}-{\varphi_0})\right]\right|\right|_2\nonumber \\
      \le \frac{||H_0^{-1}||_2}{{C_3}} 
      \left[
        \left|\left|\nabla_\varphi L_1({\varphi_0}) - \nabla_\varphi L_2({\varphi_0}) \right|\right|_2+
      \left|\left| \left[\int_0^1 [F^1_{\varphi_{t}}-F^2_{\varphi_{t}}]dt\right]({\varphi_1}-{\varphi_0})\right|\right|_2
      \right] \nonumber \\
      \le \frac{||H_0^{-1}||_2}{{C_3}} 
      \left[
        \left|\left|\nabla_\varphi L_1({\varphi_0}) - \nabla_\varphi L_2({\varphi_0}) \right|\right|_2+
       \left[\int_0^1 \left|\left|\delta F_{\varphi_{t}}\right|\right|_2dt\right]\cdot \left|\left|{\varphi_1}-{\varphi_0}\right|\right|_2
       \right]\label{eq:bound}
      \le  \frac{c}{{C_3}} [C_1+C_2R_0],
    \end{gather}   
   where $\delta F$ means $F^1-F^2$. The last inequality rises from Assumption \ref{as2} and \ref{as3}.
   The last term is a constant, which finishes the proof.
   \end{proof}
   According to \eqref{eq:bound}, 
we see that the distance between two local minimums can be bounded by 
the difference of the gradient $\nabla_\varphi L({\varphi_0})$ of two loss functions at ${\varphi_0}$,
and the difference between the two Fisher information matrices in the interpolation line.

\end{document}